\documentclass{article}

\usepackage[final,main]{neurips_2026}\usepackage{amsmath}
\usepackage{amssymb}
\usepackage{float}
\usepackage{wrapfig}
\usepackage{graphicx}
\usepackage{booktabs}
\usepackage{multirow}
\usepackage{makecell}

\newcommand{\rescell}[2]{\makecell{\textbf{#1} \\ \scriptsize #2}}

\usepackage[utf8]{inputenc} 
\usepackage[T1]{fontenc}    
\usepackage{hyperref}       
\usepackage{url}            
\usepackage{booktabs}       
\usepackage{amsfonts}       
\usepackage{nicefrac}       
\usepackage{microtype}      
\usepackage{xcolor}         
\usepackage{algorithm}
\usepackage{algpseudocode}

\title{Attention-Discounted Adaptive Sampler \\ for Masked Diffusion Language Models}

\author{%
  \begin{tabular}{c@{\qquad\qquad}c}
    Yusuf Sahin\textsuperscript{1} & Ahmed Rockey Saikia\textsuperscript{2} \\
    Volkan Cevher\textsuperscript{2} & Paolo Favaro\textsuperscript{1}
  \end{tabular} \\[0.75em]
  \textsuperscript{1}University of Bern, Bern, Switzerland \\
  \textsuperscript{2}EPFL, Lausanne, Switzerland
}

\begin{document}

\maketitle

\begin{abstract}
Masked diffusion language models can reduce inference steps by revealing multiple tokens per denoising iteration, but this parallelism is fragile: positions that are individually confident may be unsafe to commit together when their predictions are coupled. Existing training-free samplers such as Top-\(k\), Fast-dLLM, and EB-Sampler mainly control how many tokens to reveal, while often ranking candidates by token-wise scores that ignore interactions within the selected set. We propose ADAS, a training-free reranking rule that leaves the base sampler's stopping rule unchanged and greedily discounts each token-wise confidence score according to its attention to already selected positions, weighted by their prediction uncertainty. Across LLaDA-8B-Base and Dream-7B-Base on the reasoning benchmarks GSM8K and MATH500 and the code benchmarks HumanEval and MBPP, plugging ADAS into all three samplers improves low-NFE performance at matched denoiser evaluations by \(9.11\) and \(10.46\) percentage points on average, respectively, with \(3.1\%\) per-forward runtime overhead. Code is available at \url{https://github.com/yusufsahin99/ADAS}.
\end{abstract}

\section{Introduction}

Masked diffusion language models (MDLMs) \citep{sahoo2024simpleeffectivemaskeddiffusion, shi2024simplified} promise a different speed--quality tradeoff from causal autoregressive models: each denoising step can reveal a subset of positions rather than a single next token. This parallelism is especially appealing for reasoning and code-generation tasks, where long outputs make inference latency costly. The central decoding question is therefore not only \emph{what} token to place at each masked position, but also \emph{which positions} can be safely committed in the same step.

\looseness-1This second question is where aggressive parallel decoding becomes fragile. Many practical samplers rank candidate positions by token-wise quantities such as confidence, margin, or entropy, and then use a stopping rule or budget to decide how many positions to reveal. Such rules implicitly treat high-scoring candidates as compatible. Yet successful joint unmasking depends not only on individual token scores, but also on whether the selected predictions are mutually compatible: revealing one position may substantially change the predictive distribution at another.

Recent work has made this failure mode increasingly visible. Some methods treat diffusion decoding as planning or search~\citep{peng2025pathplanning,lee2025lookaheadunmaskingelicitsaccurate}; others control the amount of parallelism through schedules~\citep{luxembourg2025planspeeddilatedscheduling,israel2025acceleratingdiffusionllmsadaptive} or learned unmasking policies~\citep{jazbec2025learningunmasking,hong2025improvingdiscretediffusionunmasking,bao2025learn2pd}. The theoretical lens of \citet{benhamu2025acceleratedsamplingmaskeddiffusion} also separates \emph{model error} from \emph{joint dependence error}, clarifying that good marginal predictions are not sufficient when several positions are committed simultaneously.

We take this perspective one step further. Widely used decoders often differ in the \emph{constraint} imposed on the selected subset, but still rely on a token-wise objective for constructing that subset. Top-$k$ fixes the cardinality, Fast-dLLM~\citep{wu2025fastdllmtrainingfreeaccelerationdiffusion} uses a confidence-threshold condition, and EB-Sampler~\citep{benhamu2025acceleratedsamplingmaskeddiffusion} enforces an entropy budget. These mechanisms decide \emph{when to stop adding tokens}; they do not, by themselves, make the marginal value of a candidate depend on which other positions have already been selected.

Our proposal is to make subset construction dependency-aware while leaving the stopping rule intact. We introduce ADAS, a training-free Attention-Discounted Adaptive Sampler that greedily builds the set of positions to unmask. Starting from standard token confidence scores, ADAS discounts a remaining candidate when it attends strongly to already selected, uncertain positions. Confidence supplies an individual-value proxy, while attention supplies a directional compatibility proxy for each append decision.

This design deliberately separates ranking from budget control. ADAS does not train a planner, add a verifier, perform lookahead search, or redesign the sampler's stopping criterion. Instead, it upgrades the greedy ranking step inside existing decoders, so the same attention-discounted update can be used with fixed-cardinality, confidence-threshold, or entropy-budget rules. This places ADAS in a complementary position to learned unmasking policies~\citep{jazbec2025learningunmasking,hong2025improvingdiscretediffusionunmasking,asano2026wheretounmaskgroundtruthguidedunmaskingorder}, lookahead methods~\citep{lee2025lookaheadunmaskingelicitsaccurate}, and schedule- or verifier-based approaches~\citep{luxembourg2025planspeeddilatedscheduling,israel2025acceleratingdiffusionllmsadaptive}.

Empirically, the benefit appears exactly where the motivation predicts: the low-NFE, highly parallel regime in which token-wise ranking is most brittle. Across mathematical reasoning and code-generation benchmarks, adding the ADAS score update consistently improves Top-$k$, EB-Sampler, and Fast-dLLM at matched denoiser-evaluation budgets.

Our contributions are as follows.
\begin{itemize}
    \item We show on real decoding trajectories that attention from one masked position to another predicts the distributional change caused by revealing the attended position, and that weighting attention by the revealed token's uncertainty strengthens this association across states. Further supporting results on the relationship between attention and stochastic dependence in synthetic data are reported in the appendix.
    \item We introduce ADAS, a training-free attention-discounted greedy selector that decouples \emph{ranking} from \emph{stopping}, plugs into Top-$k$, Fast-dLLM, and EB-Sampler, and adds only $3.1\%$ runtime overhead per model forward.
    \item Across two MDLMs and four reasoning/code benchmarks, ADAS improves matched-NFE performance by $9.11$ and $10.46$ points on average.
\end{itemize}

\section{Related work}

Recent work on masked diffusion decoding has shown that generation quality depends strongly on the order and grouping of positions unmasked at each step. Several approaches therefore treat decoding as an explicit planning or policy-learning problem. \citet{peng2025pathplanning} formulate sampling as planner-guided decoding, \citet{asano2026wheretounmaskgroundtruthguidedunmaskingorder} separate where-to-unmask from what-to-unmask and learn a supervised planner, \citet{lee2025lookaheadunmaskingelicitsaccurate} use lookahead search over candidate decoding trajectories, and \citet{jazbec2025learningunmasking,hong2025improvingdiscretediffusionunmasking,bao2025learn2pd} learn unmasking policies or filters that improve over hand-designed heuristics. These methods show that token-wise confidence order is often not enough, especially outside weakly parallel regimes.

A complementary line of work improves the speed--quality tradeoff of diffusion decoding through better inference schedules or auxiliary signals. \citet{luxembourg2025planspeeddilatedscheduling} propose a structural scheduler that reduces harmful interactions through dilated grouping, while \citet{israel2025acceleratingdiffusionllmsadaptive} use an auxiliary autoregressive model to adapt the amount of parallelism during decoding. \citet{benhamu2025acceleratedsamplingmaskeddiffusion} provide a useful theoretical framing by decomposing decoding error into model error and joint dependence error, motivating adaptive multi-token unmasking rules such as EB-Sampler. \citet{kim2026klassklguidedfastinference} propose KLASS, a training-free sampler that combines token confidence with temporal KL divergence between consecutive denoising distributions to select stable tokens for unmasking.

Recent dependency-aware samplers provide especially close comparisons. PUNT~\citep{azangulov2025punt} performs approximate conditional-independence testing and removes lower-confidence tokens from conflicting groups. DAWN~\citep{luo2026dawn} extracts a dependency graph, favoring positions supported by revealed context while avoiding simultaneous updates of coupled uncertain positions. DOS~\citep{zhou2026dos} instead scores attention to already revealed tokens as positive support for masked-token updates. DAPD~\citep{kim2026dapd}, the closest method to ADAS, symmetrizes and thresholds masked-position attention into a binary dependency graph, then selects an independent set for parallel decoding; an attention edge is therefore a hard conflict.

ADAS uses the same broad source of information---self-attention---but instantiates a different decoding principle. It does not construct a binary graph, threshold attention, or impose independent-set constraints. Instead, ADAS keeps attention continuous and uses it as a marginal penalty inside greedy subset construction. Thus, ADAS does not declare two positions incompatible. A strongly coupled candidate can still be selected when its confidence gain outweighs the estimated commitment risk. This soft formulation is important because attention is an imperfect proxy for harmful dependence: not every high-attention pair should be forbidden, and not every low-attention pair is guaranteed independent.

The two methods also differ in their integration point. DAPD is a standalone dependency-aware decoder that defines its own parallel batches through graph coloring. ADAS is a sampler-agnostic reranking module: it leaves the stopping rule, budget, and admissibility condition of Top-\(k\), Fast-dLLM, or EB-Sampler unchanged, and only changes the order in which candidates are proposed. In short, DAPD uses attention to construct a graph of forbidden simultaneous updates; ADAS uses attention to compute a soft marginal discount inside existing samplers. 
\section{Preliminaries}

\subsection{Notation}

Let $\mathcal{V} = \{1,\dots,K\} \uplus \{m\}$ denote a finite vocabulary augmented with a special mask token $m$. A sequence is denoted by $x \in \mathcal{V}^d$, where $d$ is the sequence length. A position $i$ is \emph{masked} if $x_i = m$ and \emph{unmasked} otherwise. For a partially masked sequence, let $\mathcal{M} \subseteq \{1,\dots,d\}$ be the set of masked positions, and let $\bar{\mathcal{M}}$ denote its complement. We write $x^{\bar{\mathcal{M}}}$ for the observed tokens. We use $q(\cdot \mid x^{\bar{\mathcal{M}}})$ to denote the true conditional distribution over masked positions given the currently revealed context and $p_\theta(\cdot \mid x^{\bar{\mathcal{M}}})$ to denote the model's conditional distribution.

\subsection{Masked Diffusion Language Models}

Masked diffusion language models \citep{sahoo2024simpleeffectivemaskeddiffusion, shi2024simplified} generate text through iterative denoising. Starting from a highly masked sequence, the model repeatedly predicts token distributions at masked positions and progressively unmasks a subset of them until no masks remain. A decoding step therefore involves two coupled decisions: \emph{what} token to place at a selected position, and \emph{where} to unmask next.

At a decoding step, an MDLM with parameters $\theta$ predicts, for each masked position $i \in \mathcal{M}$, a conditional distribution $p_\theta(x^i \mid x^{\bar{\mathcal{M}}})$.
Many existing decoding strategies then assign each masked position a token-wise score derived from this distribution, and construct a subset $S \subseteq \mathcal{M}$ of positions to unmask according to a sampler-specific stopping rule. A common choice is the confidence score
\begin{equation}
c_i = \max_{x^i} p_\theta(x^i \mid x^{\bar{\mathcal{M}}}),
\label{eq:confidence-score}
\end{equation}

namely the probability assigned to the most likely token value at position $i$. Different samplers mainly differ in how the subset $S$ is constrained, e.g., by fixing its size, thresholding confidence, or enforcing an entropy budget.

\subsection{Model error and joint dependence error}
\label{sec:error_decomposition}

A central difficulty in parallel decoding is that the model provides token-wise conditional marginals for masked positions, while decoding multiple positions in the same step requires reasoning about their \emph{joint} behavior. As emphasized by \citet{benhamu2025acceleratedsamplingmaskeddiffusion}, the error incurred by unmasking a subset $S \subseteq \mathcal{M}$ can be decomposed into a \emph{model error} term and a \emph{joint dependence error} term:
\[
\underbrace{
\sum_{i \in S}
D_{\mathrm{KL}}\!\bigl(
q(x^i \mid x^{\bar{\mathcal{M}}}),
\, p_\theta(x^i \mid x^{\bar{\mathcal{M}}})
\bigr)
}_{\text{model error}}
\;+\;
\underbrace{
\vphantom{
\sum_{i \in S}
D_{\mathrm{KL}}\!\bigl(
q(x^i \mid x^{\bar{\mathcal{M}}}),
\, p_\theta(x^i \mid x^{\bar{\mathcal{M}}})
\bigr)
}
D_{\mathrm{KL}}\!\bigl(
q(x^{S} \mid x^{\bar{\mathcal{M}}}),
\, \textstyle\prod_{i \in S} q(x^i \mid x^{\bar{\mathcal{M}}})
\bigr)
}_{\text{joint dependence error}}
\]
The model error captures inaccuracies in the model's token-wise predictions, while the joint dependence error measures the discrepancy introduced by treating the selected positions as conditionally independent. The latter is the multi-information of the true conditional distribution \(q\) over \(S\): it is model-independent and characterizes the intrinsic coupling among positions selected for simultaneous commitment.

Since \(q\) is unobserved at decoding time, ADAS uses the denoiser self-attention \(A_{ij}\) as a model-side proxy for coupling strength. When this proxy is informative, discounting candidates that attend strongly to already selected, uncertain positions should reduce the expected dependence cost of the next parallel commitment. Section~\ref{sec:real_dependency_diagnostic} tests this premise on the target benchmarks; additional synthetic experiments validating the relationship between attention and stochastic dependence are provided in Appendix~\ref{sec:controlled_dependency_validation}.

\section{Real-data dependency diagnostic}
\label{sec:real_dependency_diagnostic}

We test whether attention reflects \emph{conditional dependence} along genuine decoding trajectories: if masked position \(j\) depends on masked position \(i\), revealing \(i\) should change the model distribution at \(j\). We generate answers for 100 examples each from GSM8K, MATH500, HumanEval, and MBPP with LLaDA-8B-Base and Dream-7B-Base using confidence Top-\(k\) decoding (\(k=8\)). A state is the partially masked output at a particular unmasking step. We probe zero-indexed steps \(0,7,15,23,31\) of the 32-step GSM8K trajectories and steps \(0,15,31,47,63\) of the 64-step trajectories for the other tasks. At each state, we choose the most-confident masked position \(i\). Before revealing it, we record final-layer attention averaged across heads, \(A_{ji}\), from each of the other seven co-selected masked positions \(j\) to \(i\), then reveal only \(i\) in a counterfactual copy. Let \(p_j^0\) and \(p_j^i\) be the distributions before and after this intervention, and \(c_j^a=\max_x p_j^a(x)\). We measure
\[
K_j=D_{\mathrm{KL}}(p_j^0\|p_j^i),\qquad
\Delta c_j=c_j^i-c_j^0,\qquad
u_{ji}=A_{ji}(1-c_i^0),\qquad
s_j=c_j^0-40u_{ji}.
\]
The baseline trajectory is unchanged, yielding 500 states and 3,500 pairs per benchmark/model; Appendix~\ref{app:real-dependency-details} gives full details.

Let \(\rho\) denote Spearman correlation over all 3,500 pairs and \(\bar\rho_s\) the mean of the 500 correlations computed separately among the seven pairs in each state. The latter holds the chosen token and partially masked output fixed. Because \(c_i^0\) is constant within a state, uncertainty gating cannot alter the statewise rank of \(A_{ji}\), but it can calibrate effects across states. A stronger dependence effect should produce a lower \(s_j\); Gain is \(\bar\rho_s(c_j^0,K_j)-\bar\rho_s(s_j,K_j)\).

\begin{table}[H]
\caption{Real-data dependency diagnostic. \(\bar\rho_s\) is the mean statewise Spearman correlation and \(\rho\) correlates all pairs. Positive Gain means the ADAS score is more negatively associated with intervention KL than confidence alone.}
\label{tab:real-dependency-main}
\centering
\small
\setlength{\tabcolsep}{3.2pt}
\begin{tabular}{llrrrrrr}
\toprule
Task & Model & $\bar\rho_s(A_{ji},K_j)$ & $\rho(A_{ji},K_j)$ & $\rho(u_{ji},K_j)$ & $\bar\rho_s(c_j^0,K_j)$ & $\bar\rho_s(s_j,K_j)$ & Gain \\
\midrule
GSM8K & Dream & .358 & .311 & .823 & .045 & -.281 & \textbf{+.325} \\
       & LLaDA & .360 & .463 & .835 & .076 & -.330 & \textbf{+.405} \\
\midrule
HumanEval & Dream & .305 & .423 & .782 & -.030 & -.209 & \textbf{+.178} \\
          & LLaDA & .389 & .544 & .841 & .100 & -.363 & \textbf{+.463} \\
\midrule
MBPP & Dream & .204 & .284 & .791 & -.114 & -.211 & \textbf{+.097} \\
     & LLaDA & .251 & .327 & .794 & -.023 & -.215 & \textbf{+.193} \\
\midrule
MATH500 & Dream & .316 & .298 & .806 & -.079 & -.271 & \textbf{+.193} \\
        & LLaDA & .436 & .545 & .815 & .116 & -.279 & \textbf{+.395} \\
\bottomrule
\end{tabular}
\end{table}

Attention is positively correlated with intervention KL in every setting (\(\bar\rho_s=.204\)--\(.436\)). Uncertainty-gated attention \(u_{ji}\) is more correlated than raw attention over all pairs (\(.782\)--\(.841\) versus \(.284\)--\(.545\)), showing improved cross-state calibration. Finally, the attention-discounted confidence score \(s_j\) is more negatively correlated with KL than the token-wise confidence \(c_j^0\) in all eight statewise comparisons (Gain \(=.097\)--\(.463\)); in all eight benchmark--model settings, the 95\% paired bootstrap interval for this improvement is strictly positive. Thus, attention is predictive of conditional dependence among masked positions, and uncertainty-weighted attention is even more predictive across decoding states.

To test whether attention reranking defers tokens that benefit from additional context, we examine its first divergence from confidence ranking in each state. We denote the confidence-ranked token it defers by \(d\) and the replacement it selects by \(r\), then reveal the common first choice \(i\). Thus, \(K_d\) and \(K_r\) measure the resulting distributional changes at the deferred and replacement tokens, while \(\Delta c_d\) and \(\Delta c_r\) measure their confidence changes. Across 1,098 states with such a divergence, deferred tokens versus replacements have mean KL \(.793\) versus \(.230\), mean confidence gain \(+.092\) versus \(+.024\), and gain confidence in \(86.8\%\) versus \(62.9\%\) of cases. Attention reranking therefore preferentially delays tokens whose distributions change more and whose confidence increases more after receiving the selected token as context, while choosing less affected replacements for immediate parallel commitment; model details appear in Appendix~\ref{app:delayed-token-stability}.
\section{Method}

\subsection{Problem statement}

Consider a masked diffusion decoding step with current set of masked positions $\mathcal{M}$. The goal is to select a subset $S \subseteq \mathcal{M}$ of positions to unmask in parallel. In most existing samplers, this selection problem is implicitly decomposed into two parts: a \emph{token-wise utility} used to rank candidate positions, and a \emph{stopping criterion} that determines when enough positions have been selected.

A common choice is to use the confidence scores $c_i$ introduced in Eq.~\eqref{eq:confidence-score} to construct $S$ so as to maximize an additive objective of the form
\begin{align}
U(S)
= \sum_{i \in S} c_i,
\end{align}
subject to a sampler-specific constraint. Different decoding rules mainly differ in this constraint. For example, Top-$k$ enforces a fixed cardinality constraint $|S| = k$, Fast-dLLM uses a confidence-based stopping condition, and EB-Sampler uses an entropy-based bound.

Our starting point is that this token-wise objective ignores dependencies among positions selected in the same step. A candidate's reliability should depend not only on its own token-wise score, but also on the uncertainty of the positions with which it is jointly committed and the strength of their interactions. We therefore retain the general subset-selection view above, but replace the purely additive token-wise objective with a dependency-aware utility that accounts for pairwise interactions.
\subsection{Attention-Discounted Adaptive Sampler (ADAS)}
\label{sec:method}

ADAS replaces token-wise candidate ranking with dependency-aware greedy subset
construction, while leaving the sampler's stopping rule unchanged. At each decoding
step, let $\mathcal M$ be the set of masked positions and let $c_i$ denote the usual
confidence score for position $i \in \mathcal M$. Let
$A \in [0,1]^{|\mathcal M|\times|\mathcal M|}$ be an attention-based interaction
matrix over masked positions, where $A_{ij}$ measures attention from position $i$
to position $j$. In our implementation, $A$ is obtained from the denoiser's final
self-attention layer by averaging attention scores across heads, following the
ablation in Appendix~\ref{app:attention-source}.

Given an already selected set $S$, ADAS scores a remaining candidate $i \in \mathcal{M}\setminus S$ by the
marginal utility
\begin{align}
u(i \mid S,A)
=
c_i
-
\alpha \sum_{s\in S} A_{is}(1-c_s),
\label{eq:adas-marginal}
\end{align}
where $\alpha \ge 0$ controls the strength of the attention discount. The first term
favors positions with high token-wise confidence. The second term discounts
candidates that attend strongly to selected positions whose predictions are still
uncertain. Thus, a candidate is penalized not simply for being coupled to another
selected position, but for being coupled to one whose commitment is less reliable.

For an ordered sequence of selections \(\pi=(\pi_1,\ldots,\pi_m)\), define the append-only utility
\begin{align}
U(\pi,A)
=
\sum_{t=1}^{m}\left[
c_{\pi_t}-\alpha\sum_{r<t}A_{\pi_t\pi_r}(1-c_{\pi_r})
\right].
\label{eq:adas-ordered-utility}
\end{align}
Appending candidate \(i\) to \(\pi\) has marginal value exactly
Eq.~\eqref{eq:adas-marginal}. Equation~\eqref{eq:adas-ordered-utility} is therefore
an ordered motivation for the operational greedy rule, not an equivalent unordered
set utility: an unordered pairwise objective would introduce reverse-direction terms
absent from ADAS. Algorithm~\ref{alg:adas} defines the method; we claim no global
optimization or approximation guarantee.

At each iteration, ADAS selects the remaining candidate with largest current
marginal utility and checks whether adding it satisfies the base sampler's
admissibility rule. If the enlarged set is admissible, the candidate is added;
otherwise, construction stops and the selected positions are unmasked in parallel.

\begin{algorithm}[t]
\caption{Attention-Discounted Adaptive Sampler (ADAS)}
\label{alg:adas}
\begin{algorithmic}
\algrenewcommand\alglinenumber[1]{}
\Require $\mathcal{M}$, scores $\{c_i\}$, attention $A$, weight $\alpha$, admissibility rule $\mathrm{Adm}$
\State $S \gets \emptyset$
\While{$\mathcal{M} \setminus S \neq \emptyset$}
    \State $i^\star \gets \arg\max_{i \in \mathcal{M}\setminus S}
    \left[c_i - \alpha \sum_{s \in S} A_{is}(1-c_s)\right]$
    \If{\textbf{not} $\mathrm{Adm}(S \cup \{i^\star\})$}
        \State \textbf{break}
    \EndIf
    \State $S \gets S \cup \{i^\star\}$
\EndWhile
\State \Return $S$
\end{algorithmic}
\end{algorithm}

In our experiments, ADAS is combined with three admissibility rules:
\begin{align}
\text{Top-}k: \quad & |S \cup \{i^\star\}| \le k, \\
\text{Fast-dLLM}: \quad &
|S \cup \{i^\star\}|
\Bigl(1-\min_{s\in S\cup\{i^\star\}} c_s\Bigr) < f, \\
\text{EB-Sampler}: \quad &
\sum_{s\in S\cup\{i^\star\}}
H\!\bigl(p_\theta(x^s \mid x^{\bar{\mathcal M}})\bigr)
-\max_{s\in S\cup\{i^\star\}}
H\!\bigl(p_\theta(x^s \mid x^{\bar{\mathcal M}})\bigr)
< \gamma.
\end{align}
Here $k$, $f$, and $\gamma$ are the fixed-cardinality, confidence, and entropy-budget
thresholds, respectively. These rules determine when subset construction stops;
ADAS only changes how candidates are ranked before that stopping decision.

We use \(\alpha=40\) as a single global default in every main experiment. It was selected once by averaging an ablation over six LLaDA-8B HumanEval settings and then reused without task-, model-, or stopping-rule-specific retuning (Appendix~\ref{app:alpha-ablation}). The attention discount can be updated incrementally. After adding a selected
position $s$, each remaining candidate's score is decreased by
$\alpha A_{is}(1-c_s)$, giving $O(|S|\,|\mathcal M|)$ selection cost for a final
subset $S$. This overhead is small relative to denoiser forward passes; we measure
it in Section~\ref{sec:runtime_overhead} and ablate $\alpha$ and the uncertainty
term in Appendix~\ref{app:alpha-ablation}.

\section{Experiments}

\subsection{Experimental setup}
We evaluate ADAS on four benchmarks spanning mathematical reasoning and code generation: GSM8K \citep{cobbe2021trainingverifierssolvemath}, MATH500 \citep{hendrycks2021measuringmathematicalproblemsolving}, MBPP \citep{austin2021programsynthesislargelanguage}, and HumanEval \citep{chen2021evaluatinglargelanguagemodels}. We use two strong open-weight masked diffusion language models, Dream-7B-Base \citep{ye2025dream7bdiffusionlarge} and LLaDA-8B-Base \citep{nie2025largelanguagediffusionmodels}.

Our goal is to test whether attention-discounted subset construction helps precisely where parallel unmasking is most useful but most error-prone. We therefore instantiate ADAS with three representative stopping criteria: Top-\(k\), Fast-dLLM, and EB-Sampler, yielding three variants we denote Top-\(k\)+AD, Fast-dLLM+AD, and EB+AD. Each variant is compared against its corresponding baseline over a range of parallel decoding settings.

We further evaluate KLASS \citep{kim2026klassklguidedfastinference} and DAPD \citep{kim2026dapd} as additional training-free baselines.

\subsection{Main results}
\label{sec:experiments_main_results}

\begin{figure*}[t]
    \centering
    \includegraphics[width=\textwidth]{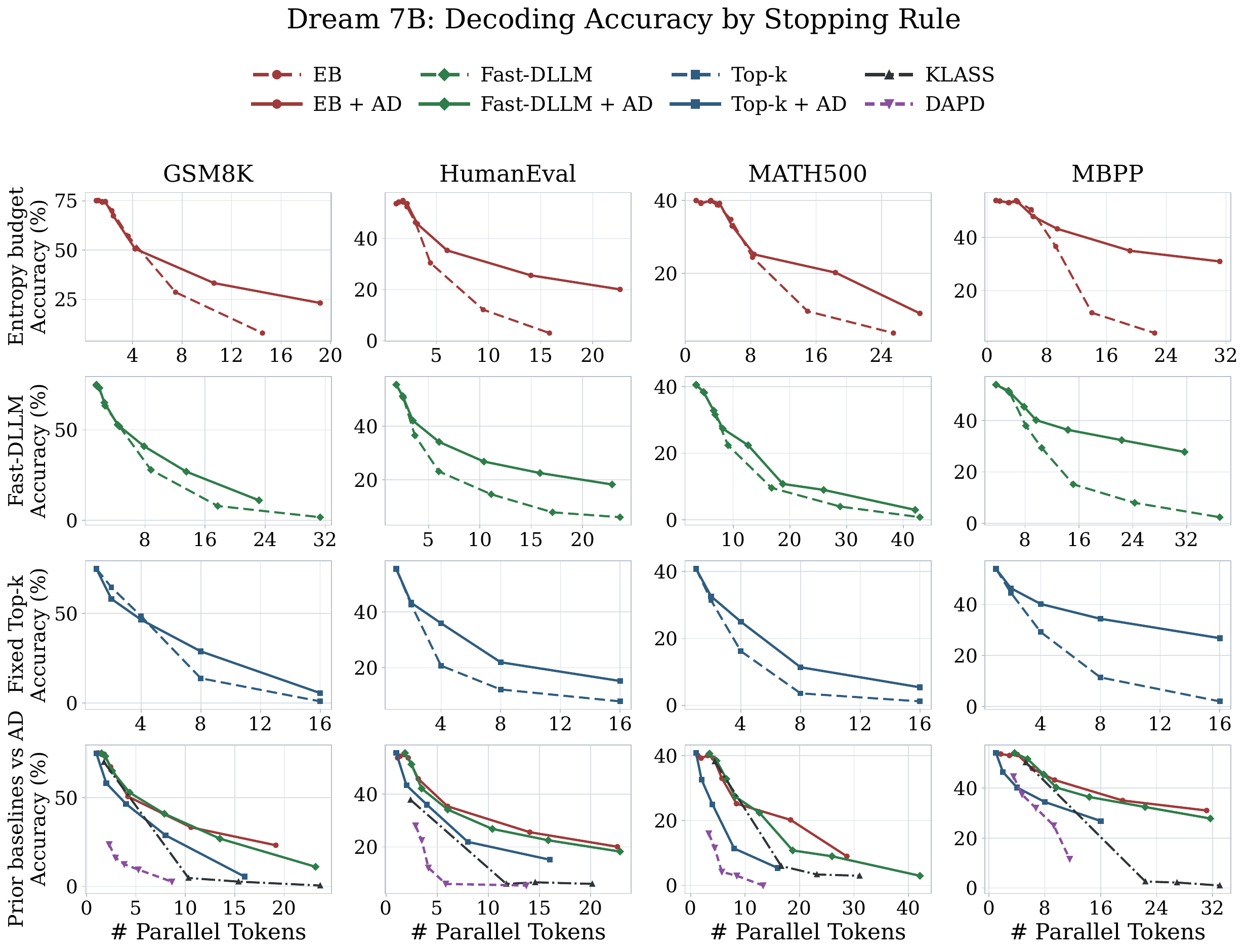}
    \caption{Effect of attention-discounted selection on Dream-7B-Base. Rows correspond to
    entropy-bounded decoding, Fast-dLLM, fixed Top-\(k\), and comparison against KLASS and DAPD;
    columns correspond to GSM8K, HumanEval, MATH500, and MBPP. The same pattern observed
    for LLaDA-8B-Base persists: attention-discounted selection is most beneficial when many
    tokens are revealed in parallel, where the original samplers degrade sharply.}
    \label{fig:dream_main_results}
\end{figure*}

\begin{figure*}[t]
    \centering
    \includegraphics[width=\textwidth]{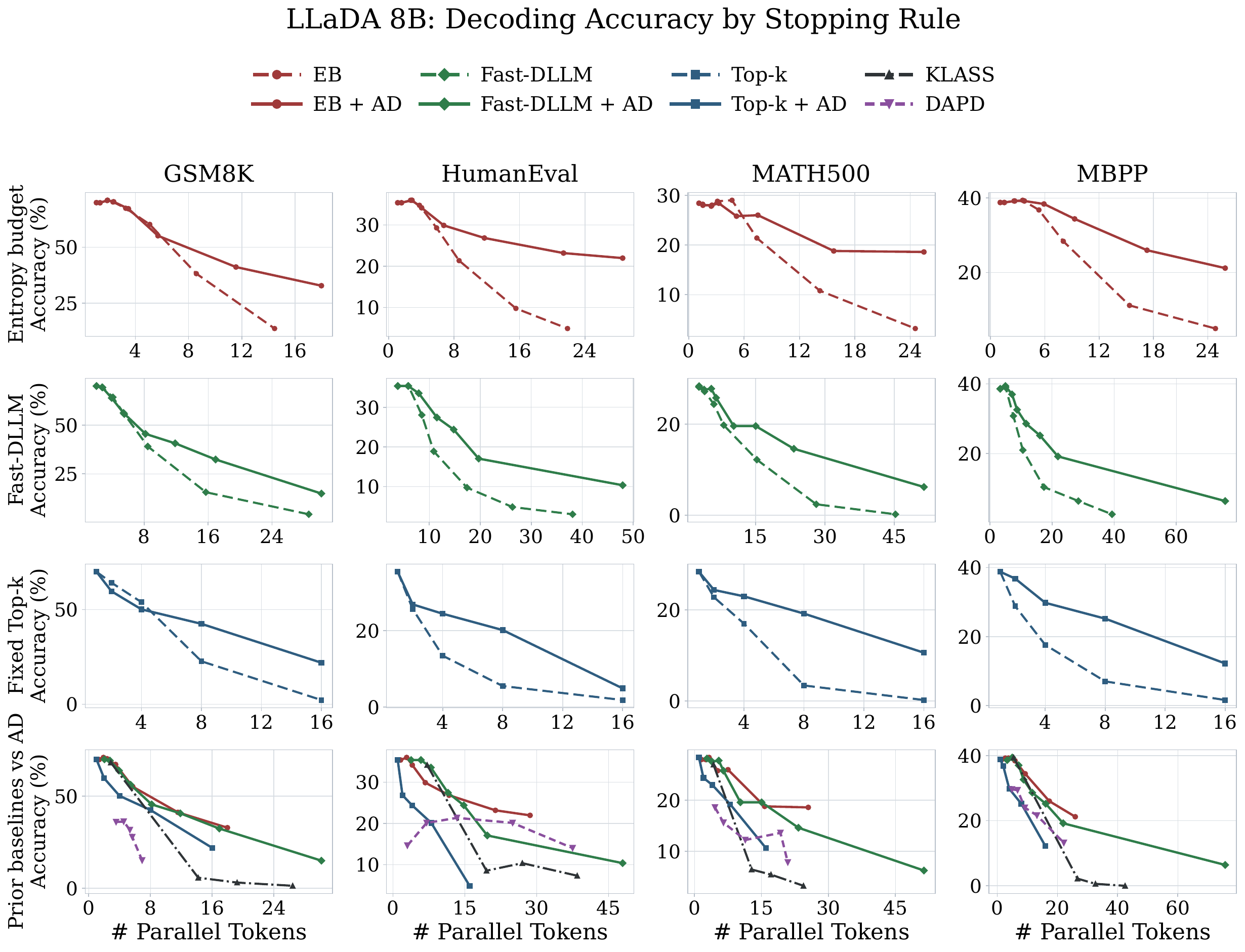}
    \caption{Effect of attention-discounted selection on LLaDA-8B-Base. Rows correspond to
    entropy-bounded decoding, Fast-dLLM, fixed Top-\(k\), and comparison against KLASS and DAPD;
    columns correspond to GSM8K, HumanEval, MATH500, and MBPP. Across stopping rules and
    datasets, attention-discounted selection preserves substantially higher accuracy in the
    highly parallel regime, while matching the original samplers more closely when the number
    of parallel tokens is small.}
    \label{fig:llada_main_results}
\end{figure*}

Figures~\ref{fig:dream_main_results} and \ref{fig:llada_main_results}  summarize the main speed--quality tradeoff. Moving to the right corresponds to revealing more tokens per denoising step, and therefore to lower NFE. The central question is whether the sampler can preserve quality in this highly parallel regime, where jointly selected dependent positions are most likely to induce errors.

In each figure, the first three rows compare EB-Sampler, Fast-dLLM, and Top-\(k\) with their ADAS variants; the final row compares ADAS-augmented samplers with KLASS and DAPD. Together, these plots test two claims: ADAS should improve its own base sampler, and the resulting decoders should remain competitive with recent training-free samplers.


Across both models, ADAS consistently improves robustness in the high-parallelism regime.
For the adaptive stopping rules EB and fast-dLLM, the baseline curves often deteriorate rapidly as the entropy
budget increases, whereas their attention-discounted variants degrade more gradually.

Fixed Top-\(k\) decoding shows that the benefit is not tied to adaptive stopping criteria. Since
\(k\) directly controls the number of tokens unmasked at each step, this setting isolates the effect
of subset construction. Across both base models, Top-\(k\)+AD improves or preserves performance
at larger \(k\), where vanilla Top-\(k\) suffers the largest degradation. Thus, reranking candidates
by attention-discounted utility selects safer groups of tokens even when the subset size is fixed
in advance.

The comparison with prior training-free baselines further supports this interpretation.
In the bottom row of each figure, the ADAS-augmented decoders generally remain competitive
with or above KLASS and DAPD at moderate and large parallel-token counts. Both baselines
often degrade rapidly as parallelism increases, whereas ADAS tends to degrade more smoothly.
One possible explanation for the KLASS trend is that temporal-stability signals become less
reliable when many tokens are committed at once, since consecutive denoising distributions
can shift substantially. The DAPD trend suggests that hard attention-derived constraints may
be brittle in some high-parallel regimes. Overall, these results indicate that soft
attention-discounted reranking is a robust way to reduce harmful joint commitments among
dependent positions.

To quantify the low-NFE regime, for baseline points averaging at least four tokens per step, we interpolate the ADAS curve at matched NFE. Table~\ref{tab:matched-nfe-gains-exact-llada-dream} reports mean gains across the three stopping rules as LLaDA / Dream.

\begin{table}[!t]
\caption{Matched-NFE absolute gains in the low-NFE regime, reported as LLaDA-8B-Base / Dream-7B-Base percentage points.}
\label{tab:matched-nfe-gains-exact-llada-dream}
\centering
\small
\begin{tabular}{lcccc}
\toprule
Dataset & EB & Fast-dLLM & Top-\(k\) & Avg. over methods \\
\midrule
GSM8K & +9.22 / +14.31 & +6.92 / +6.36 & +11.90 / +5.88 & +9.35 / +8.85 \\
MATH500 & +6.49 / +4.84 & +5.70 / +2.15 & +10.73 / +6.93 & +7.64 / +4.64 \\
HumanEval & +10.45 / +15.63 & +6.90 / +12.03 & +9.55 / +10.77 & +8.97 / +12.81 \\
MBPP & +10.37 / +15.01 & +7.39 / +12.04 & +13.67 / +19.60 & +10.48 / +15.55 \\
\midrule
Avg. over datasets & +9.13 / +12.45 & +6.73 / +8.14 & +11.46 / +10.80 & +9.11 / +10.46 \\
\bottomrule
\end{tabular}
\end{table}

The gains hold across entropy-, confidence-, and fixed-budget stopping rules and are strongest on code generation, where simultaneous coupled commitments can create inconsistent outputs. The smaller but positive MATH500 gains suggest that longer sequential chains, rather than pairwise incompatibilities within one update, may govern some errors.

Across \(90\) matched operating points, ADAS has mean gain \(+9.27\) with a 95\% bootstrap interval of \([+7.74,+10.84]\), improving in \(80\) cases and regressing in \(10\). This measures cross-operating-point stability, not fixed-setting significance; Appendix~\ref{app:robustness-operating-points} gives the full breakdown.

\subsection{Analysis of runtime overhead}
\label{sec:runtime_overhead}

ADAS reads last-layer attention and greedily updates token scores; accessing these weights requires disabling FlashAttention only in the final denoiser layer. Comparing each variant with its vanilla decoder under the same stopping rule and similar NFE, full-dataset wall-clock measurements show that attention extraction and reranking add $3.1\%$ per model forward on average.

\section{Discussion and limitations}
\label{sec:discussion}

ADAS is a training-free, dependency-aware sampler that improves highly parallel decoding across models, tasks, and stopping rules. Its central limitation is the proxy itself: attention may reflect syntax, position, or formatting, omits higher-order interactions, and weak attention does not guarantee compatibility. Although our intervention validates conditional influence on all four target benchmarks, evidence remains limited to two base MDLMs and reasoning/code; head- or layer-specific signals, larger models, and open-ended generation remain untested. The global \(\alpha=40\) was chosen on one model/benchmark, and confidence miscalibration can directly affect its uncertainty gate.

ADAS also commits irreversibly. Revise-or-remask could repair early errors after neighboring context arrives, but would change commitment dynamics and possibly NFE, making it complementary future work. Deterministic temperature-zero runs have no seed variance: the operating-point bootstrap measures cross-configuration stability, not fixed-setting significance, and sensitivity to resampled few-shot orderings remains untested. Higher-order signals, revision, and auxiliary verification are promising extensions.

The broader impacts of ADAS are mostly inherited from the underlying language models. ADAS may improve the inference efficiency and output quality of existing masked diffusion language models, which can reduce decoding cost and make these models more practical to use. At the same time, faster or higher-quality generation could indirectly increase misuse risks already associated with language models, such as automated spam, low-quality content generation, or other harmful downstream applications. ADAS does not introduce a new model, dataset, or deployed system, so these risks are inherited from the underlying models rather than specific to a newly released artifact.

\bibliographystyle{plainnat}
\bibliography{references}

\appendix
\section{Evaluation details}
\label{app:eval-details}

We evaluate on GSM8K, HumanEval, MBPP, and MATH500 with Dream 7B and LLaDA 8B. We use lm-eval-harness 0.4.8 with its default few-shot seed 1234: 8-shot prompting for GSM8K, 4-shot for MATH500, 3-shot for MBPP, and 0-shot for HumanEval. For batched inference, we pad each prompt to the task sequence length. Maximum generation length is $256$ for GSM8K and $512$ for HumanEval, MBPP, and MATH500. All evaluations use temperature $0$. Dream uses Transformers 4.57.1 with flash SDPA disabled when attention extraction is required; its prompts and one-example generation were checked against the literal harness request.

\paragraph{Decoding controls.}
For the entropy-budget (EB) and Fast-dLLM families, we sweep the budget parameters $\gamma$ and $f$. For the fixed-subset Top-$k$ family, we sweep the subset size $k$ directly.

\begin{table}[h]
\caption{Hyperparameter ranges used for the three decoder families.}
\label{tab:eval-sweeps}
\centering
\small
\begin{tabular}{l l}
\hline
Method family & Sweep values \\
\hline
EB / EB+AD & $\gamma \in \{10^{-3}, 10^{-2}, 10^{-1}, 0.3, 1, 3, 10, 20\}$ \\
Fast-dLLM & $ f \in \{0.3, 1, 2, 3, 5, 8, 12\}$ \\
Fast-dLLM+AD & $ f \in \{0.3, 1, 2, 3, 5, 8, 12, 20\}$ \\
Top-$k$ / Top-$k$+AD & $k \in \{1,2,4,8,16\}$ \\
\hline
\end{tabular}

\end{table}
For EB and Fast-dLLM, larger threshold permits more aggressive parallel unmasking. For Top-$k$, the level of parallelism is controlled directly by $k$, i.e., the number of tokens unmasked per denoising step. We include the extra \(f=20\) point for Fast-dLLM+AD only to probe the very high-parallel regime reached by the vanilla decoder at smaller thresholds; matched-NFE summaries use interpolation at baseline operating points rather than threshold-wise comparisons.
\paragraph{Compute resources.}
All evaluations were run as batched inference jobs on one 4 $\times$ GH200 GPU node per operating point. We used padding to the task maximum
sequence length for batched inference. Dream 7B used batch size 4 per GPU, and LLaDA 8B used batch size
2 per GPU.

\begin{table}[h]
\caption{Actual wall-clock duration by model and dataset, aggregated over the completed full-dataset operating points used in the paper figures.}
\label{tab:eval-compute-by-task}
\centering
\small
\begin{tabular}{llr}
\toprule
Model & Dataset & Min--max hrs/run \\
\midrule
Dream 7B & GSM8K & 0.13--2.21 \\
Dream 7B & HumanEval & 0.06--0.57 \\
Dream 7B & MATH500 & 0.08--2.00 \\
Dream 7B & MBPP & 0.10--2.15 \\
\midrule
LLaDA 8B & GSM8K & 0.26--4.86 \\
LLaDA 8B & HumanEval & 0.07--1.26 \\
LLaDA 8B & MATH500 & 0.14--3.67 \\
LLaDA 8B & MBPP & 0.15--4.27 \\
\bottomrule
\end{tabular}
\end{table}

\paragraph{Hyperparameters for KLASS and DAPD.}
For KLASS and DAPD, we did not perform per-task accuracy tuning. Instead, we first ran small pilot evaluations to identify threshold values that place each method in the same broad low-NFE regime as our main curves, and then reused the same hyperparameter grids for all models and datasets. For KLASS, which accepts tokens satisfying both a confidence and KL criterion, we evaluated
\[
(\theta_{\mathrm{conf}}, \theta_{\mathrm{KL}})
\in
\{(0.70,1.00), (0.198,0.51), (0.18,0.60), (0.16,0.70)\}.
\]
For DAPD, using the masked max-normalized attention dependency score, we evaluated
\[
(\tau_{\min}, \tau_{\max})
\in
\{(0.002,0.02), (0.005,0.05), (0.01,0.10), (0.02,0.20), (0.05,0.50)\}.
\]
These settings were chosen to span a range of parallelism levels rather than to maximize accuracy on any individual benchmark.

\section{Real-data dependency diagnostic details}
\label{app:real-dependency-details}

\paragraph{Trajectory construction.}
For every example and model, we run the ordinary confidence Top-\(k\) trajectory with \(k=8\) and collect the partially masked outputs at the five unmasking steps listed below. MATH500 uses a 100-example subset balanced across its seven subjects (15 examples each from algebra and counting/probability and 14 from each remaining subject).

\begin{table}[h]
\caption{Settings for the real-data intervention diagnostic.}
\centering
\small
\begin{tabular}{lrrrrl}
\toprule
Task & Examples & Shots & Length & Steps & Probed steps \\
\midrule
GSM8K & 100 & 8 & 256 & 32 & 0, 7, 15, 23, 31 \\
MATH500 & 100 & 4 & 512 & 64 & 0, 15, 31, 47, 63 \\
HumanEval & 100 & 0 & 512 & 64 & 0, 15, 31, 47, 63 \\
MBPP & 100 & 3 & 512 & 64 & 0, 15, 31, 47, 63 \\
\bottomrule
\end{tabular}
\end{table}

At each state, let \(M\) be the masked positions and \(p_m^{\mathrm{before}}\) the first-pass predictive distribution. We choose \(i=\arg\max_{m\in M}c_m\), where \(c_m=\max_v p_m^{\mathrm{before}}(v)\), and set its deterministic prediction to \(\hat x_i=\arg\max_v p_i^{\mathrm{before}}(v)\). From this same pre-reveal pass, we extract head-mean final-layer \(A_{ji}\), with query \(j\) and key \(i\). In a non-mutating counterfactual copy, we reveal only \(i\), rerun the model, and record
\[
D_{\mathrm{KL}}\!\left(p_j^{\mathrm{before}}\middle\|p_j^{\mathrm{after}}\right),
\qquad
\Delta c_j=\max_v p_j^{\mathrm{after}}(v)-\max_v p_j^{\mathrm{before}}(v).
\]
The original trajectory then commits its ordinary Top-\(k\) set. The primary scope contains the other seven co-selected positions, giving 3,500 pairs per task/model. This scope directly measures dependencies among positions the baseline would commit together.

\paragraph{Estimands and uncertainty.}
The correlation \(\rho\) combines all 3,500 pairs. To avoid conflating token ranking with differences across examples and denoising stages, we also compute Spearman correlation independently within each state and report their mean \(\bar\rho_s\). Intervals for these means are normal intervals across states. For paired comparisons, we first average the five states per example and bootstrap the 100 example-level means. Since \(i\) and \(c_i\) are fixed within a state, \(A_{ji}\) and \(A_{ji}(1-c_i)\) have identical statewise ranks; gating is evaluated as cross-state calibration through \(\rho\).

\begin{table}[h]
\caption{Statewise intervention results with uncertainty intervals. Gain is \(\bar\rho_s(c_j^0,K_j)-\bar\rho_s(s_j,K_j)\). Gain intervals use a paired bootstrap over 100 examples.}
\label{tab:real-dependency-intervals}
\centering
\small
\setlength{\tabcolsep}{5pt}
\begin{tabular}{llcc}
\toprule
Task & Model & \(\bar\rho_s(A_{ji},K_j)\) [95\% interval] & Gain [95\% interval] \\
\midrule
GSM8K & Dream & .358 [.312, .403] & .325 [.295, .358] \\
       & LLaDA & .360 [.312, .409] & .405 [.368, .444] \\
HumanEval & Dream & .305 [.265, .346] & .178 [.143, .216] \\
          & LLaDA & .389 [.340, .437] & .463 [.402, .525] \\
MBPP & Dream & .204 [.161, .247] & .097 [.074, .123] \\
     & LLaDA & .251 [.202, .301] & .193 [.165, .221] \\
MATH500 & Dream & .316 [.274, .359] & .193 [.164, .223] \\
        & LLaDA & .436 [.393, .480] & .395 [.347, .445] \\
\bottomrule
\end{tabular}
\end{table}

\paragraph{Delayed-token stability.}
\label{app:delayed-token-stability}
After the common first token \(i\), we compare the next candidate under confidence Top-8 with the exact Top-8+AD rule. In 1,098 states they differ, defining exactly one deferred token and one ADAS replacement per state. We analyze this first divergence, not all eight selections, and measure both tokens after revealing \(i\).

\begin{table}[h]
\caption{Change after receiving newly revealed context, reported as deferred token / ADAS replacement.}
\centering
\small
\begin{tabular}{lccc}
\toprule
Model & KL change & Mean \(\Delta c_j\) & Confidence increased \\
\midrule
LLaDA & .671 / .197 & +.084 / +.035 & 86.2\% / 64.5\% \\
Dream & .914 / .263 & +.101 / +.013 & 87.4\% / 61.2\% \\
\textbf{Overall} & \textbf{.793 / .230} & \textbf{+.092 / +.024} & \textbf{86.8\% / 62.9\%} \\
\bottomrule
\end{tabular}
\end{table}

The confidence-gain direction holds in all eight task/model settings, while KL is higher for deferred tokens in seven of eight (Dream/MBPP is effectively tied). Because free-form solutions do not provide a unique token-aligned target, this diagnostic measures stability rather than token correctness; correctness is measured by end-to-end task scores.

\section{Theoretical motivation for the ADAS utility}
\label{app:theoretical_grounding}

We provide a local derivation motivating the ADAS utility. The argument is not meant
to establish a global guarantee for an arbitrary transformer; rather, it shows that
under a standard first-order approximation of the denoiser around the current decoding
state, the natural dependence penalty has the same form as the ADAS discount.

Fix the current partially unmasked sequence \(x^{\bar{\mathcal M}}\). For each masked
position \(s\), let
\[
p_s(a)=p_\theta(x^s=a\mid x^{\bar{\mathcal M}}),\qquad
\hat x_s=\arg\max_{a\in\mathcal V}p_s(a),\qquad
c_s=p_s(\hat x_s).
\]
Let \(Z_s=\phi_s(X^s)\) be the representation inserted at position \(s\), and let
\(\hat z_s=\phi_s(\hat x_s)\). Since the vocabulary is finite, define
\[
D_\phi^2
=
\max_{s\in\mathcal M}
\max_{a,b\in\mathcal V}
\|\phi_s(a)-\phi_s(b)\|_2^2 .
\]
Then
\[
\begin{aligned}
\mathbb E\!\left[\|Z_s-\hat z_s\|_2^2\right]
&=
\sum_{a\in\mathcal V}p_s(a)
\|\phi_s(a)-\phi_s(\hat x_s)\|_2^2  \\
&\le
D_\phi^2
\sum_{a\ne \hat x_s}p_s(a) \\
&=
D_\phi^2(1-c_s).
\end{aligned}
\]
Thus \(1-c_s\) upper-bounds the representation uncertainty of the selected source
token \(s\).

Now consider appending a candidate position \(i\) after an already selected set \(S\).
Let \(\ell_i(z_S)\) denote the logits at position \(i\) when the selected positions
are filled with representations \(z_S=\{z_s:s\in S\}\). Around the current decoding
state \(\hat z_S\), a first-order expansion gives
\[
\ell_i(z_S)
\approx
\ell_i(\hat z_S)
+
\sum_{s\in S}
J_{is}(z_s-\hat z_s),
\]
where \(J_{is}\) is the Jacobian of the logits at position \(i\) with respect to the
representation at position \(s\).

For a self-attention layer, the direct message from source \(s\) to target \(i\) is
weighted by the attention coefficient \(A_{is}\). We therefore use the empirical
attention matrix as a tractable proxy for the relative sensitivity of \(i\) to
perturbations at \(s\), and write the local approximation
\[
\|J_{is}\|_2^2
\approx
L_i^2 A_{is},
\]
where \(L_i\) absorbs the remaining local linear maps, residual transformations, and
normalization factors. This approximation says that source positions receiving larger
attention from \(i\) have larger local influence on the logits at \(i\).

Using this approximation,
\[
\begin{aligned}
\mathbb E\!\left[
\|\ell_i(Z_S)-\ell_i(\hat z_S)\|_2^2
\right]
&\approx
\mathbb E\!\left[
\left\|
\sum_{s\in S}
J_{is}(Z_s-\hat z_s)
\right\|_2^2
\right] \\
&\lesssim
L_i^2
\sum_{s\in S}
A_{is}
\mathbb E\!\left[
\|Z_s-\hat z_s\|_2^2
\right] \\
&\le
L_i^2D_\phi^2
\sum_{s\in S}
A_{is}(1-c_s).
\end{aligned}
\]
Finally, for softmax probabilities, local logit perturbations induce KL changes that
are second order in the perturbation size:
\[
D_{\mathrm{KL}}
\left(
\operatorname{softmax}(\ell_i+\delta)
\;\middle\|\;
\operatorname{softmax}(\ell_i)
\right)
\le
\frac12\|\delta\|_2^2 .
\]
Therefore, the incremental dependence risk of adding candidate \(i\) after \(S\) is
locally controlled by a quantity proportional to
\[
\sum_{s\in S}A_{is}(1-c_s).
\]

This motivates the risk-adjusted marginal utility
\[
u(i\mid S,A)
=
c_i
-
\alpha
\sum_{s\in S}
A_{is}(1-c_s),
\]
where \(\alpha\ge 0\) absorbs the constants in the local approximation and controls
the strength of the dependence penalty.
\section{Hyperparameter and ablation studies}
\label{app:alpha-ablation}

\paragraph{Attention-discount strength}
We ablate the attention-discount strength $\alpha$ on LLaDA-8B HumanEval under three decoding families: fixed Top-$k$, AD+EB, and Fast-dLLM+AD. For fixed Top-$k$, we test $k \in \{8,16\}$. For AD+EB, we use two entropy-budget settings chosen to match these operating regimes approximately: $\gamma=10$ (low NFE) and $\gamma=3$ (medium NFE). For AD Fast-dLLM, we similarly use $f=5$ and $f=2$. In all cases, we sweep
\[
\alpha \in \{1,5,10,20,40,80\},
\]
and compare against the corresponding no-AD baseline, which we denote as $\alpha=0$.

\begin{table}[h]
\caption{Pass@1 accuracy (\%) for the $\alpha$ sweep on LLaDA HumanEval across fixed Top-$k$, AD+EB, and AD Fast-dLLM. The no-AD baseline is shown as $\alpha=0$.}
\label{tab:alpha-ablation}
\centering
\small
\begin{tabular}{l c c c c c c c}
\toprule
Setting & $\alpha=0$ & $\alpha=1$ & $\alpha=5$ & $\alpha=10$ & $\alpha=20$ & $\alpha=40$ & $\alpha=80$ \\
\midrule
Top-$k$, $k=16$ & 1.83 & 4.88 & 7.32 & 5.49 & 5.49 & 4.88 & 4.88 \\
Top-$k$, $k=8$  & 5.49 & 9.15 & 18.90 & 18.29 & 21.34 & 20.12 & 18.90 \\
AD+EB, $\gamma=10$ & 9.76 & 9.15 & 20.12 & 23.78 & 23.78 & 23.17 & 23.78 \\
AD+EB, $\gamma=3$  & 21.34 & 20.73 & 23.17 & 22.56 & 26.22 & 26.83 & 22.56 \\
Fast-dLLM+AD, $f=5$ & 9.76 & 9.15 & 21.95 & 23.78 & 24.39 & 24.39 & 21.95 \\
Fast-dLLM+AD, $f=2$ & 28.05 & 29.27 & 27.44 & 28.05 & 30.49 & 33.54 & 33.54 \\
\midrule
Average over 6 settings & -- & 13.72 & 19.82 & 20.33 & 21.95 & \textbf{22.15} & 20.93 \\
\bottomrule
\end{tabular}
\end{table}

The best $\alpha$ depends on the decoding regime when viewed per setting. For example, fixed Top-$k$ prefers smaller values in some cases, whereas the adaptive stopping-rule methods are more favorable to larger values. However, when averaging across all six matched settings in Table~\ref{tab:alpha-ablation}, $\alpha=40$ achieves the highest overall accuracy. Relative to an oracle that selects the best $\alpha$ separately for each row, the global default loses only $0.72$ points on average and at most $2.44$ points in the worst case. We therefore use $\alpha=40$ as a single global default in the main experiments, avoiding task- or regime-specific retuning while remaining competitive across the tested settings.

\paragraph{Ablation on the uncertainty weighting term.}
We additionally ablate the uncertainty factor used inside the attention discount.
The default ADAS utility discounts a candidate position according to the uncertainty
of the already selected position, using the factor \(1-c_s\). We compare this choice
against a variant that replaces \(1-c_s\) with a global average uncertainty over the
currently masked positions, denoted \(1-c_{\mathrm{mask}}\). This ablation is conducted
on LLaDA-8B-Base with HumanEval under Top-\(k\)+AD decoding at \(k=8\).

\begin{table}[h]
\centering
\caption{Ablation of the uncertainty weighting term in the attention discount on
LLaDA-8B-Base HumanEval with Top-\(k\)+AD decoding at \(k=8\). The default
selected-token uncertainty \(1-c_s\) outperforms using the average uncertainty over
masked positions \(1-c_{\mathrm{mask}}\).}
\label{tab:uncertainty_weight_ablation}
\begin{tabular}{lc}
\toprule
Uncertainty weighting term & Pass@1 (\%) \\
\midrule
Selected-token uncertainty \(1-c_s\) & \textbf{20.12} \\
Average masked-position uncertainty \(1-c_{\mathrm{mask}}\) & 17.07 \\
\bottomrule
\end{tabular}
\end{table}

The selected-token uncertainty weighting improves Pass@1 by \(3.05\) absolute points
over the global masked-position uncertainty variant. This suggests that the discount
benefits from conditioning on the uncertainty of the specific selected position rather
than using a single global uncertainty scalar.

\paragraph{Effect of attention source}
\label{app:attention-source}

We ablate which layer is used to construct the attention matrix in ADAS. We use LLaDA-8B-Base on HumanEval with fixed Top-$k$ decoding at $k=8$, so all variants use the same number of denoising steps. The no-ADAS baseline ranks tokens only by confidence, while the ADAS variants use mean attention over heads from the specified transformer layer.

\begin{table}[h]
\caption{Attention-source ablation for ADAS on LLaDA-8B HumanEval with fixed Top-$k$ decoding at $k=8$. Final-layer attention performs best, suggesting that later attention patterns provide the most useful dependency signal for selecting tokens to unmask in parallel.}
\label{tab:attention-source-ablation}
\centering
\small
\begin{tabular}{lc}
\toprule
Selection rule & Pass@1 (\%) \\
\midrule
Top-$k$ confidence, no ADAS & 5.49 \\
Top-$k$+AD with first-layer attention & 9.76 \\
Top-$k$+AD with middle-layer attention & 15.24 \\
Top-$k$+AD with final-layer attention &\textbf{ 20.12} \\
\bottomrule
\end{tabular}
\end{table}

\section{Robustness across operating points}
\label{app:robustness-operating-points}

We perform an additional robustness analysis over matched operating points. Each operating point corresponds to a model--dataset--stopping-rule--hyperparameter configuration. This analysis is intended to assess whether the observed gains are stable across the evaluated settings, rather than driven by a small number of favorable thresholds or tasks.

For each vanilla baseline point in the low-NFE regime, defined as having at least four parallel tokens per denoising step on average, we linearly interpolate the corresponding ADAS curve at the same NFE and compute the matched gain
\[
\Delta = \mathrm{score}_{\mathrm{ADAS}}(\mathrm{NFE}) -
         \mathrm{score}_{\mathrm{base}}(\mathrm{NFE}).
\]
Baseline points outside the interpolation range of the corresponding ADAS curve are excluded. We then bootstrap the matched gains over operating points with 10{,}000 resamples and report the mean gain and central 95\% bootstrap interval. Because operating points from the same model, dataset, or decoding family are correlated, this is a descriptive stability summary across evaluated configurations, not a confidence interval for a fixed setting or a paired per-example significance test.

\begin{table}[h]
\caption{Robustness of ADAS gains across matched operating points. Gains are absolute accuracy/pass@1 differences in percentage points after matching by NFE. Bootstrap intervals use 10{,}000 operating-point resamples and summarize cross-configuration stability.}
\label{tab:robustness-by-stopping-rule}
\centering
\small
\begin{tabular}{lcccc}
\toprule
Method family & Matched points & Mean gain & \makecell{95\% bootstrap\\interval} & Positive / negative \\
\midrule
EB-Sampler
& 28 & +10.42 & [+7.30, +13.67] & 24 / 4 \\
Fast-dLLM
& 38 & +7.24 & [+5.42, +9.14] & 34 / 4 \\
Top-$k$
& 24 & +11.13 & [+8.31, +13.97] & 22 / 2 \\
\midrule
All
& 90 & +9.27 & [+7.74, +10.84] & 80 / 10 \\
\bottomrule
\end{tabular}
\end{table}

\begin{table}[h]
\caption{Robustness of ADAS gains across datasets under the same matched-operating-point analysis as Table~\ref{tab:robustness-by-stopping-rule}.}
\label{tab:robustness-by-dataset}
\centering
\small
\begin{tabular}{lcccc}
\toprule
Dataset & Matched points & Mean gain & \makecell{95\% bootstrap\\interval} & Positive / negative \\
\midrule
GSM8K
& 19 & +8.60 & [+4.71, +12.48] & 14 / 5 \\
HumanEval
& 22 & +10.33 & [+8.10, +12.50] & 21 / 1 \\
MATH500
& 24 & +5.73 & [+3.84, +7.64] & 21 / 3 \\
MBPP
& 25 & +12.24 & [+8.99, +15.59] & 24 / 1 \\
\bottomrule
\end{tabular}
\end{table}
\section{Controlled synthetic dependency diagnostics}
\label{sec:controlled_dependency_validation}

We complement the real-data intervention with arithmetic predicates whose dependency structure is known. For the first diagnostic, each example contains three sum or product predicates and exactly two masked numeric literals. In the dependent condition, both masks occur in the same predicate; in the independent condition, they occur in different predicates. Dream-7B-Base predicts both literals in one forward pass without confidence ordering, adaptive stopping, or attention discounting. Under constraint-satisfaction scoring, dependent pairs are substantially harder: accuracy falls from \(71\%\) for independent pairs to \(30\%\) for dependent pairs.

\begin{figure}[h]
    \centering
    \includegraphics[width=0.58\textwidth]{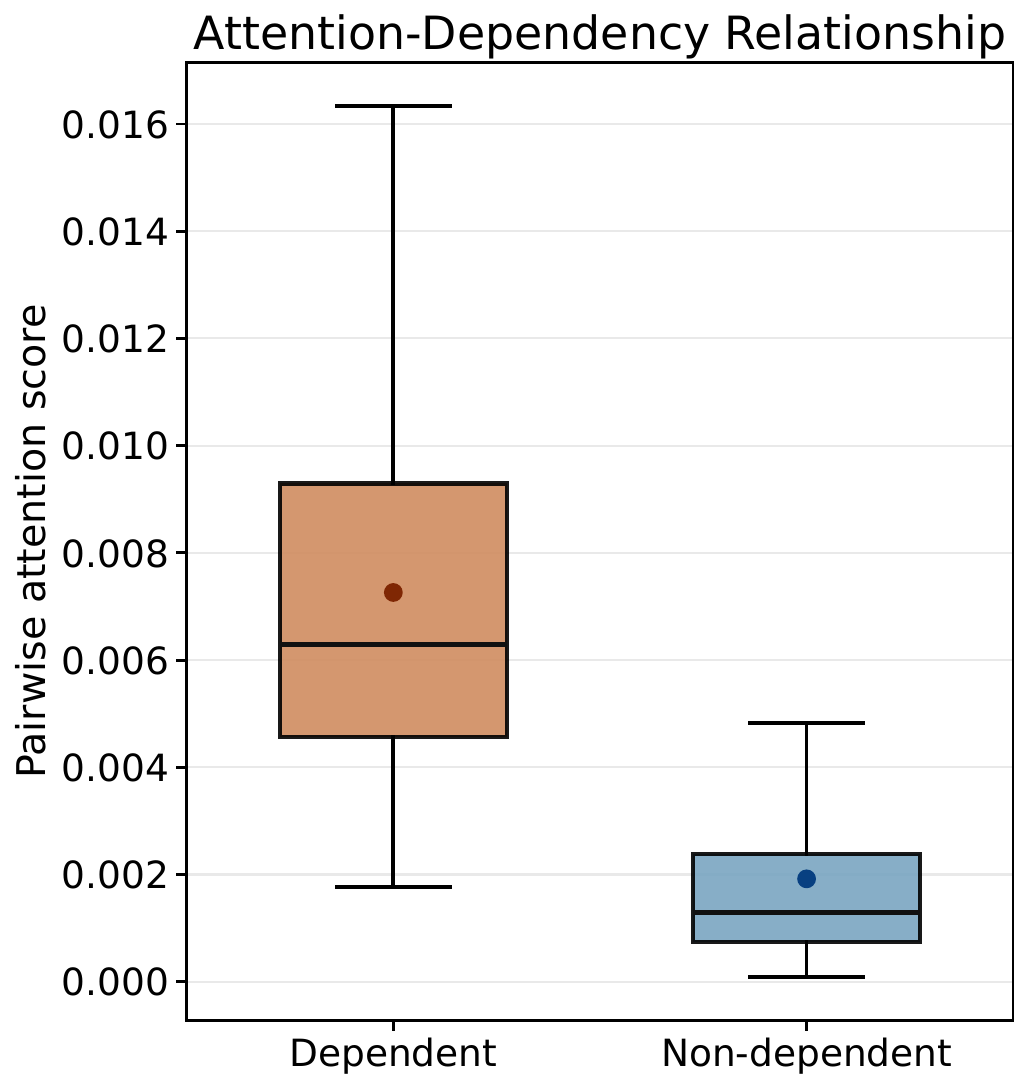}
    \caption{Pairwise attention for dependent and non-dependent masked-token pairs in the synthetic diagnostic on LLaDA-8B-Base.}
    \label{fig:attention-dependency-boxplot}
\end{figure}

The second diagnostic includes sum, product, minimum, and maximum predicates. We compare final-layer, head-mean attention between masked tokens from the same predicate with pairs from different predicates. Figure~\ref{fig:attention-dependency-boxplot} and Table~\ref{tab:toy_attention_dependency_shared_k1_compact} show higher attention for known-dependent pairs on both models. This controlled result isolates dependency structure, while Section~\ref{sec:real_dependency_diagnostic} establishes the corresponding intervention relationship on the target tasks.

\begin{figure}[H]
\centering
\setlength{\fboxsep}{8pt}
\fbox{%
\begin{minipage}{0.94\linewidth}
\small

\textbf{Dataset Construction}

\medskip
Assume the set of equation types
\[
E =
\left\{
A + B + C = D,\;
A \cdot B \cdot C = D,\;
\min\{A,B,C\} = D,\;
\max\{A,B,C\} = D
\right\}
\]
is given, together with the number of predicates $n$ and masking ratio $r$.

\medskip
\textbf{1. Sample predicate indices.}
We sample a sequence
\[
\pi = (\pi_1,\ldots,\pi_n), \qquad \pi_i \in \{0,\ldots,|E|-1\},
\]
and construct the $i$-th predicate from the equation type $E[\pi_i]$.
\medskip\\
\textbf{2. Instantiate the predicates.}
Each predicate contains four integer slots $(A,B,C,D)$, so the full sequence contains $4n$ integer variables. We sample positive integers for $A,B,C$ and compute $D$ according to the selected equation type.
\medskip\\
\textbf{3. Mask integer variables.}
We randomly mask
\[
r \cdot 4n
\]
of the integer positions in the sequence. Only the integers are masked; the operators and relation symbols ($+$, $\cdot$, $\min$, $\max$, $=$) are never masked.
\end{minipage}%
}
\caption{Illustration of the toy data-generation process. The construction yields a masked sequence containing masked-token pairs with known dependency structure.}
\label{fig:toy-dependency-construction}
\end{figure}

\begin{table}[h]
\centering
\small
\caption{
Attention-dependency relationship on the shared toy dataset.
Reported values are mean pairwise attention for dependent and non-dependent masked pairs aggregated as mean of attention heads from the last layer.
}
\label{tab:toy_attention_dependency_shared_k1_compact}
\begin{tabular}{lcc}
\hline
Model & Dep. Attn. & Non-Dep. Attn. \\
\hline
LLaDA-8B-Base & 0.007261 & 0.001919 \\
Dream-7B-Base & 0.009517 & 0.004842 \\
\hline
\end{tabular}
\end{table}

\section{Detailed results}
\begin{table*}[t]
\centering
\small
\setlength{\tabcolsep}{6pt}
\renewcommand{\arraystretch}{1.2}
\caption{
 Results on \textbf{Dream-7B-Base} with EB stopping rule. Each cell reports \textbf{score} on the first line and average NFE over samples on the second line. For GSM8K and MATH500, score is accuracy; for HumanEval and MBPP, score is Pass@1.
}
\label{tab:eb_dream_appendix}
\begin{tabular}{c cc|cc|cc|cc}
\toprule
& \multicolumn{2}{c}{GSM8K} 
& \multicolumn{2}{c}{HumanEval}
& \multicolumn{2}{c}{MATH500}
& \multicolumn{2}{c}{MBPP} \\
\cmidrule(lr){2-3} \cmidrule(lr){4-5} \cmidrule(lr){6-7} \cmidrule(lr){8-9}
$\gamma$ & EB & EB + AD & EB & EB + AD & EB & EB + AD & EB & EB + AD \\
\midrule
0.001 & \rescell{75.06}{240.50} & \rescell{75.06}{240.37}
      & \rescell{53.66}{453.44} & \rescell{53.66}{453.06}
      & \rescell{40.00}{386.65} & \rescell{40.00}{384.99}
      & \rescell{53.80}{432.49} & \rescell{53.80}{431.56} \\

\midrule
0.01  & \rescell{75.13}{209.23} & \rescell{75.13}{208.93}
      & \rescell{54.27}{372.90} & \rescell{54.27}{372.83}
      & \rescell{39.40}{269.48} & \rescell{39.20}{266.06}
      & \rescell{53.60}{305.49} & \rescell{53.60}{302.70} \\

\midrule
0.1   & \rescell{74.60}{166.87} & \rescell{74.30}{165.76}
      & \rescell{54.27}{289.84} & \rescell{54.88}{287.95}
      & \rescell{39.80}{168.08} & \rescell{40.00}{163.52}
      & \rescell{53.00}{179.38} & \rescell{53.00}{174.84} \\

\midrule
0.3   & \rescell{74.68}{143.42} & \rescell{74.45}{142.59}
      & \rescell{52.44}{237.53} & \rescell{53.66}{236.08}
      & \rescell{38.80}{131.00} & \rescell{39.20}{122.17}
      & \rescell{53.80}{131.58} & \rescell{53.40}{126.62} \\

\midrule
1     & \rescell{69.98}{110.42} & \rescell{67.32}{104.99}
      & \rescell{46.34}{172.41} & \rescell{45.73}{162.49}
      & \rescell{34.80}{91.65}  & \rescell{33.00}{89.23}
      & \rescell{50.40}{86.70}  & \rescell{47.80}{82.90} \\

\midrule
3     & \rescell{57.09}{70.41}  & \rescell{50.64}{61.00}
      & \rescell{30.49}{116.12} & \rescell{35.37}{85.06}
      & \rescell{24.40}{62.27}  & \rescell{25.20}{60.81}
      & \rescell{36.60}{55.71}  & \rescell{43.20}{54.47} \\

\midrule
10    & \rescell{28.66}{34.29}  & \rescell{33.28}{24.21}
      & \rescell{12.20}{54.01}  & \rescell{25.61}{36.43}
      & \rescell{9.60}{34.24}   & \rescell{20.20}{27.88}
      & \rescell{11.80}{36.50}  & \rescell{35.00}{26.77} \\

\midrule
20    & \rescell{7.96}{17.66}   & \rescell{23.20}{13.36}
      & \rescell{3.05}{32.32}   & \rescell{20.12}{22.63}
      & \rescell{3.60}{20.12}   & \rescell{9.00}{17.85}
      & \rescell{4.20}{22.82}   & \rescell{31.00}{16.43} \\
\bottomrule
\end{tabular}
\end{table*}
\begin{table*}[t]
\centering
\small
\setlength{\tabcolsep}{6pt}
\renewcommand{\arraystretch}{1.2}
\caption{
 Results on \textbf{LLaDA-8B-Base} with EB stopping rule. Each cell reports \textbf{score} on the first line and average NFE over samples on the second line. For GSM8K and MATH500, score is accuracy; for HumanEval and MBPP, score is Pass@1.
}
\label{tab:eb_llada_appendix}
\begin{tabular}{c cc|cc|cc|cc}
\toprule
& \multicolumn{2}{c}{GSM8K} 
& \multicolumn{2}{c}{HumanEval}
& \multicolumn{2}{c}{MATH500}
& \multicolumn{2}{c}{MBPP} \\
\cmidrule(lr){2-3} \cmidrule(lr){4-5} \cmidrule(lr){6-7} \cmidrule(lr){8-9}
$\gamma$ & EB & EB + AD & EB & EB + AD & EB & EB + AD & EB & EB + AD \\
\midrule
0.001 & \rescell{69.90}{237.17} & \rescell{69.90}{236.92}
      & \rescell{35.37}{457.12} & \rescell{35.37}{456.04}
      & \rescell{28.40}{460.24} & \rescell{28.40}{459.09}
      & \rescell{38.80}{475.93} & \rescell{38.80}{475.47} \\

\midrule
0.01  & \rescell{69.83}{190.83} & \rescell{69.83}{189.83}
      & \rescell{35.37}{326.08} & \rescell{35.37}{321.84}
      & \rescell{28.20}{336.50} & \rescell{28.00}{332.56}
      & \rescell{38.80}{338.33} & \rescell{38.80}{334.52} \\

\midrule
0.1   & \rescell{70.81}{136.13} & \rescell{70.96}{134.03}
      & \rescell{35.98}{190.35} & \rescell{35.98}{178.60}
      & \rescell{27.80}{209.92} & \rescell{28.00}{203.39}
      & \rescell{39.20}{197.32} & \rescell{39.20}{191.63} \\

\midrule
0.3   & \rescell{70.20}{110.54} & \rescell{70.28}{107.93}
      & \rescell{34.76}{135.54} & \rescell{34.15}{126.72}
      & \rescell{28.80}{164.38} & \rescell{28.40}{157.04}
      & \rescell{39.40}{144.04} & \rescell{39.20}{136.83} \\

\midrule
1     & \rescell{67.40}{78.10}  & \rescell{67.17}{73.56}
      & \rescell{29.27}{87.19}  & \rescell{29.88}{75.84}
      & \rescell{29.00}{109.11} & \rescell{25.80}{98.88}
      & \rescell{36.80}{95.70}  & \rescell{38.40}{86.68} \\

\midrule
3     & \rescell{60.20}{50.36}  & \rescell{55.12}{44.89}
      & \rescell{21.34}{59.38}  & \rescell{26.83}{43.71}
      & \rescell{21.40}{69.33}  & \rescell{26.00}{68.29}
      & \rescell{28.40}{63.66}  & \rescell{34.40}{55.08} \\

\midrule
10    & \rescell{38.21}{29.82}  & \rescell{41.17}{22.14}
      & \rescell{9.76}{32.90}   & \rescell{23.17}{23.95}
      & \rescell{10.80}{36.02}  & \rescell{18.80}{32.59}
      & \rescell{11.20}{33.34}  & \rescell{26.00}{29.61} \\

\midrule
20    & \rescell{13.72}{17.70}  & \rescell{32.83}{14.24}
      & \rescell{4.88}{23.41}   & \rescell{21.95}{17.90}
      & \rescell{3.20}{20.86}   & \rescell{18.60}{20.09}
      & \rescell{5.00}{20.60}   & \rescell{21.20}{19.75} \\
\bottomrule
\end{tabular}
\end{table*}
\begin{table*}[t]
\centering
\small
\setlength{\tabcolsep}{6pt}
\renewcommand{\arraystretch}{1.2}
\caption{
 Results on \textbf{Dream-7B-Base} with Fast-dLLM stopping rule. Each cell reports \textbf{score} on the first line and average NFE over samples on the second line. For GSM8K and MATH500, score is accuracy; for HumanEval and MBPP, score is Pass@1.
}
\label{tab:fastdllm_dream_appendix}
\begin{tabular}{c cc|cc|cc|cc}
\toprule
& \multicolumn{2}{c}{GSM8K} 
& \multicolumn{2}{c}{HumanEval}
& \multicolumn{2}{c}{MATH500}
& \multicolumn{2}{c}{MBPP} \\
\cmidrule(lr){2-3} \cmidrule(lr){4-5} \cmidrule(lr){6-7} \cmidrule(lr){8-9}
$f$ & FD & FD + AD & FD & FD + AD & FD & FD + AD &FD & FD + AD \\
\midrule
0.3 & \rescell{74.45}{167.07} & \rescell{74.98}{167.59}
    & \rescell{55.49}{273.27} & \rescell{55.49}{275.57}
    & \rescell{40.40}{146.80} & \rescell{40.60}{148.19}
    & \rescell{54.00}{138.66} & \rescell{54.00}{139.72} \\

\midrule
1   & \rescell{73.01}{132.59} & \rescell{73.31}{134.97}
    & \rescell{50.61}{201.44} & \rescell{51.22}{205.78}
    & \rescell{38.20}{104.11} & \rescell{38.40}{107.50}
    & \rescell{51.00}{90.36}  & \rescell{51.60}{92.76} \\

\midrule
2   & \rescell{63.31}{95.36}  & \rescell{65.05}{98.78}
    & \rescell{36.59}{139.48} & \rescell{42.07}{147.07}
    & \rescell{31.60}{75.19}  & \rescell{32.80}{78.06}
    & \rescell{38.00}{63.20}  & \rescell{45.40}{65.30} \\

\midrule
3   & \rescell{51.86}{55.60}  & \rescell{52.84}{58.71}
    & \rescell{23.17}{85.32}  & \rescell{34.15}{84.87}
    & \rescell{22.40}{56.29}  & \rescell{27.40}{62.73}
    & \rescell{29.40}{48.96}  & \rescell{40.20}{53.23} \\

\midrule
5   & \rescell{27.90}{29.12}  & \rescell{40.94}{32.49}
    & \rescell{14.63}{46.10}  & \rescell{26.83}{49.23}
    & \rescell{9.60}{30.52}   & \rescell{22.40}{40.51}
    & \rescell{15.20}{33.78}  & \rescell{36.40}{35.64} \\

\midrule
8   & \rescell{7.96}{14.47}   & \rescell{26.91}{18.96}
    & \rescell{7.93}{29.98}   & \rescell{22.56}{32.28}
    & \rescell{4.00}{17.73}   & \rescell{10.80}{27.31}
    & \rescell{8.00}{21.09}   & \rescell{32.40}{22.91} \\

\midrule
12  & \rescell{1.74}{8.17}    & \rescell{11.07}{11.04}
    & \rescell{6.10}{21.65}   & \rescell{18.29}{22.38}
    & \rescell{0.80}{11.92}   & \rescell{9.00}{19.72}
    & \rescell{2.40}{13.87}   & \rescell{27.80}{16.16} \\
\bottomrule
\end{tabular}
\end{table*}
\begin{table*}[t]
\centering
\small
\setlength{\tabcolsep}{6pt}
\renewcommand{\arraystretch}{1.2}
\caption{
 Results on \textbf{LLaDA-8B-Base} with Fast-dLLM stopping rule. Each cell reports \textbf{score} on the first line and average NFE over samples on the second line. For GSM8K and MATH500, score is accuracy; for HumanEval and MBPP, score is Pass@1.
}
\label{tab:fastdllm_llada_appendix}
\begin{tabular}{c cc|cc|cc|cc}
\toprule
& \multicolumn{2}{c}{GSM8K} 
& \multicolumn{2}{c}{HumanEval}
& \multicolumn{2}{c}{MATH500}
& \multicolumn{2}{c}{MBPP} \\
\cmidrule(lr){2-3} \cmidrule(lr){4-5} \cmidrule(lr){6-7} \cmidrule(lr){8-9}
$f$ & FD & FD + AD & FD & FD + AD & FD & FD + AD &FD & FD + AD \\
\midrule
0.3 & \rescell{69.98}{125.16} & \rescell{69.98}{126.02}
    & \rescell{35.37}{132.63} & \rescell{35.37}{134.80}
    & \rescell{28.40}{182.54} & \rescell{28.20}{184.13}
    & \rescell{38.60}{153.71} & \rescell{38.60}{154.44} \\

\midrule
1   & \rescell{69.22}{91.28}  & \rescell{69.29}{93.16}
    & \rescell{35.37}{86.65}  & \rescell{35.37}{87.59}
    & \rescell{27.20}{128.43} & \rescell{27.60}{131.62}
    & \rescell{39.00}{101.49} & \rescell{39.40}{102.71} \\

\midrule
2   & \rescell{64.22}{62.60}  & \rescell{63.84}{65.10}
    & \rescell{28.05}{60.01}  & \rescell{33.54}{64.59}
    & \rescell{24.40}{85.34}  & \rescell{27.80}{94.18}
    & \rescell{30.80}{67.83}  & \rescell{37.00}{71.91} \\

\midrule
3   & \rescell{55.57}{46.39}  & \rescell{56.18}{47.26}
    & \rescell{18.90}{47.09}  & \rescell{27.44}{44.57}
    & \rescell{19.80}{63.04}  & \rescell{25.80}{79.03}
    & \rescell{21.00}{48.29}  & \rescell{32.60}{58.54} \\

\midrule
5   & \rescell{39.04}{30.29}  & \rescell{45.49}{31.37}
    & \rescell{9.76}{29.43}   & \rescell{24.39}{34.63}
    & \rescell{12.20}{33.48}  & \rescell{19.60}{49.76}
    & \rescell{10.40}{29.45}  & \rescell{28.60}{43.96} \\

\midrule
8   & \rescell{15.62}{16.29}  & \rescell{40.64}{21.56}
    & \rescell{4.88}{19.46}   & \rescell{17.07}{26.01}
    & \rescell{2.40}{18.20}   & \rescell{19.60}{34.05}
    & \rescell{6.40}{17.98}   & \rescell{25.20}{31.80} \\

\midrule
12  & \rescell{4.32}{8.96}    & \rescell{32.37}{15.12}
    & \rescell{3.05}{13.44}   & \rescell{10.37}{10.68}
    & \rescell{0.20}{11.32}   & \rescell{14.60}{21.99}
    & \rescell{2.60}{13.03}   & \rescell{19.20}{23.37} \\

\midrule
20  & \multicolumn{2}{c|}{}
    & \multicolumn{2}{c|}{}
    & \rescell{}{ } & \rescell{6.20}{9.96}
    & \multicolumn{1}{c}{}
    & \rescell{6.40}{6.76} \\
\bottomrule
\end{tabular}
\end{table*}
\begin{table*}[t]
\centering
\small
\setlength{\tabcolsep}{6pt}
\renewcommand{\arraystretch}{1.2}
\caption{
 Results on \textbf{Dream-7B-Base} with Top-$k$ stopping rule. Each cell reports \textbf{score} on the first line and average NFE over samples on the second line. For GSM8K and MATH500, score is accuracy; for HumanEval and MBPP, score is Pass@1.
}
\label{tab:topk_dream_appendix}
\begin{tabular}{c cc|cc|cc|cc}
\toprule
& \multicolumn{2}{c}{GSM8K} 
& \multicolumn{2}{c}{HumanEval}
& \multicolumn{2}{c}{MATH500}
& \multicolumn{2}{c}{MBPP} \\
\cmidrule(lr){2-3} \cmidrule(lr){4-5} \cmidrule(lr){6-7} \cmidrule(lr){8-9}
$k$ & Top-k & Top-k + AD & Top-k & Top-k + AD & Top-k & Top-k + AD & Top-k & Top-k + AD \\
\midrule
1  & \rescell{74.91}{256.00} & \rescell{74.91}{256.00}
   & \rescell{55.49}{512.00} & \rescell{55.49}{512.00}
   & \rescell{40.80}{512.00} & \rescell{40.80}{512.00}
   & \rescell{54.00}{512.00} & \rescell{54.00}{512.00} \\

\midrule
2  & \rescell{64.67}{128.00} & \rescell{58.15}{128.00}
   & \rescell{42.68}{256.00} & \rescell{43.29}{256.00}
   & \rescell{31.40}{256.00} & \rescell{32.60}{256.00}
   & \rescell{44.60}{256.00} & \rescell{46.40}{256.00} \\

\midrule
4  & \rescell{48.45}{64.00}  & \rescell{46.47}{64.00}
   & \rescell{20.73}{128.00} & \rescell{35.98}{128.00}
   & \rescell{16.20}{128.00} & \rescell{25.00}{128.00}
   & \rescell{29.20}{128.00} & \rescell{40.20}{128.00} \\

\midrule
8  & \rescell{13.80}{32.00}  & \rescell{28.81}{32.00}
   & \rescell{12.20}{64.00}  & \rescell{21.95}{64.00}
   & \rescell{3.60}{64.00}   & \rescell{11.40}{64.00}
   & \rescell{11.40}{64.00}  & \rescell{34.40}{64.00} \\

\midrule
16 & \rescell{0.91}{16.00}   & \rescell{5.53}{16.00}
   & \rescell{7.93}{32.00}   & \rescell{15.24}{32.00}
   & \rescell{1.20}{32.00}   & \rescell{5.40}{32.00}
   & \rescell{2.00}{32.00}   & \rescell{26.80}{32.00} \\
\bottomrule
\end{tabular}
\end{table*}
\begin{table*}[t]
\centering
\small
\setlength{\tabcolsep}{6pt}
\renewcommand{\arraystretch}{1.2}
\caption{
 Results on \textbf{LLaDA-8B-Base} with Top-$k$ stopping rule. Each cell reports \textbf{score} on the first line and average NFE over samples on the second line. For GSM8K and MATH500, score is accuracy; for HumanEval and MBPP, score is Pass@1.
}
\label{tab:topk_llada_appendix}
\begin{tabular}{c cc|cc|cc|cc}
\toprule
& \multicolumn{2}{c}{GSM8K} 
& \multicolumn{2}{c}{HumanEval}
& \multicolumn{2}{c}{MATH500}
& \multicolumn{2}{c}{MBPP} \\
\cmidrule(lr){2-3} \cmidrule(lr){4-5} \cmidrule(lr){6-7} \cmidrule(lr){8-9}
$k$ & Top-k & Top-k + AD & Top-k & Top-k + AD & Top-k & Top-k + AD & Top-k & Top-k + AD \\
\midrule
1  & \rescell{69.98}{256.00} & \rescell{69.98}{256.00}
   & \rescell{35.37}{512.00} & \rescell{35.37}{512.00}
   & \rescell{28.40}{512.00} & \rescell{28.40}{512.00}
   & \rescell{38.80}{512.00} & \rescell{38.80}{512.00} \\

\midrule
2  & \rescell{64.22}{128.00} & \rescell{59.67}{128.00}
   & \rescell{25.61}{256.00} & \rescell{26.83}{256.00}
   & \rescell{22.80}{256.00} & \rescell{24.40}{256.00}
   & \rescell{28.80}{256.00} & \rescell{36.80}{256.00} \\

\midrule
4  & \rescell{53.90}{64.00}  & \rescell{50.11}{64.00}
   & \rescell{13.41}{128.00} & \rescell{24.39}{128.00}
   & \rescell{17.00}{128.00} & \rescell{23.00}{128.00}
   & \rescell{17.60}{128.00} & \rescell{29.80}{128.00} \\

\midrule
8  & \rescell{22.67}{32.00}  & \rescell{42.46}{32.00}
   & \rescell{5.49}{64.00}   & \rescell{20.12}{64.00}
   & \rescell{3.40}{64.00}   & \rescell{19.20}{64.00}
   & \rescell{7.00}{64.00}   & \rescell{25.20}{64.00} \\

\midrule
16 & \rescell{2.20}{16.00}   & \rescell{21.91}{16.00}
   & \rescell{1.83}{32.00}   & \rescell{4.88}{32.00}
   & \rescell{0.20}{32.00}   & \rescell{10.60}{32.00}
   & \rescell{1.60}{32.00}   & \rescell{12.20}{32.00} \\
\bottomrule
\end{tabular}
\end{table*}

\clearpage
\section*{NeurIPS Paper Checklist}

\begin{enumerate}

\item {\bf Claims}
    \item[] Question: Do the main claims made in the abstract and introduction accurately reflect the paper's contributions and scope?
    \item[] Answer: \answerYes{} 
    \item[] Justification: The main claims are stated in the abstract and introduction and are supported by the method description in Section~\ref{sec:method}, the main experimental results in Section~\ref{sec:experiments_main_results}, and the runtime analysis in Section~\ref{sec:runtime_overhead}. The scope is limited to masked diffusion language models, the evaluated benchmarks, and the stopping criteria considered in the experiments.

\item {\bf Limitations}
    \item[] Question: Does the paper discuss the limitations of the work performed by the authors?
    \item[] Answer: \answerYes{} 
    \item[] Justification: The limitations are discussed in Section~\ref{sec:discussion}, including the use of self-attention only as a proxy for stochastic dependence, the limited model and task scope, and the fact that robustness intervals summarize variation across matched operating points rather than repeated seeds or per-example uncertainty.

\item {\bf Theory assumptions and proofs}
    \item[] Question: For each theoretical result, does the paper provide the full set of assumptions and a complete (and correct) proof?
    \item[] Answer: \answerNA{} 
    \item[] Justification: The paper does not present formal theoretical results, theorems, or proofs. The error decomposition in Section~\ref{sec:error_decomposition} is used as motivation from prior work, while the proposed ADAS method is evaluated empirically in Section~\ref{sec:experiments_main_results}.

    \item {\bf Experimental result reproducibility}
    \item[] Question: Does the paper fully disclose all the information needed to reproduce the main experimental results of the paper to the extent that it affects the main claims and/or conclusions of the paper (regardless of whether the code and data are provided or not)?
    \item[] Answer: \answerYes{} 
    \item[] Justification: We provide a detailed description of the evaluation settings and proposed method in the paper, and include in the supplemental material a link to a public repository containing the code, commands, and instructions needed to reproduce the main experiments. The implementation builds on the standard language model evaluation framework \texttt{lm-eval}.

\item {\bf Open access to data and code}
    \item[] Question: Does the paper provide open access to the data and code, with sufficient instructions to faithfully reproduce the main experimental results, as described in supplemental material?
    \item[] Answer: \answerYes{} 
    \item[] Justification: We provide reproduction code and instructions in the supplemental material.

\item {\bf Experimental setting/details}
    \item[] Question: Does the paper specify all the training and test details (e.g., data splits, hyperparameters, how they were chosen, type of optimizer) necessary to understand the results?
    \item[] Answer: \answerYes{} 
    \item[] Justification: The proposed ADAS decoding method is described in Section~\ref{sec:method}, including the attention-based interaction matrix, marginal utility, greedy subset construction, stopping-rule-agnostic formulation, and computational complexity. Algorithm~\ref{alg:adas} gives pseudocode, and Appendix~\ref{app:eval-details} specifies the experimental methodology, including benchmarks, models, prompting setup, decoding sweep values, batch sizes, and compute resources.

\item {\bf Experiment statistical significance}
    \item[] Question: Does the paper report error bars suitably and correctly defined or other appropriate information about the statistical significance of the experiments?
    \item[] Answer: \answerYes{} 
    \item[] Justification: The paper reports a robustness analysis over matched low-NFE operating points in Appendix ~\ref{app:robustness-operating-points}. Specifically, for each baseline operating point, we interpolate the corresponding ADAS curve at the same NFE and compute the matched accuracy/pass@1 gain. We then bootstrap these matched gains over operating points using 10,000 resamples and report the mean gain, 95\% bootstrap interval, and the number of positive and negative matched comparisons for each stopping-rule family. These intervals are intended as a robustness summary across evaluated model--dataset--sampler configurations rather than as repeated-seed uncertainty estimates.

\item {\bf Experiments compute resources}
    \item[] Question: For each experiment, does the paper provide sufficient information on the computer resources (type of compute workers, memory, time of execution) needed to reproduce the experiments?
    \item[] Answer: \answerYes{} 
    \item[] Justification: We provide extensive information regarding the compute resources in Section \ref{app:eval-details}.
    
\item {\bf Code of ethics}
    \item[] Question: Does the research conducted in the paper conform, in every respect, with the NeurIPS Code of Ethics \url{https://neurips.cc/public/EthicsGuidelines}?
    \item[] Answer: \answerYes{} 
    \item[] Justification: The research complies with the NeurIPS Code of Ethics: it does not involve human subjects, crowdsourcing, private or newly collected personal data, surveillance data, or high-risk deployment. Experiments use standard public benchmarks and existing open-weight models, and the main ethical considerations are indirect risks from improving language-model generation speed and quality, which are discussed as inherited from downstream use of the underlying models.

\item {\bf Broader impacts}
    \item[] Question: Does the paper discuss both potential positive societal impacts and negative societal impacts of the work performed?
    \item[] Answer: \answerYes{} 
    \item[] Justification: Section~\ref{sec:discussion} discusses broader impacts. ADAS may improve the inference efficiency and output quality of existing masked diffusion language models, reducing decoding cost and making these models more practical to use. The paper also discusses potential negative impacts: faster or higher-quality generation could indirectly increase misuse risks already associated with language models, such as automated spam, low-quality content generation, or other harmful downstream applications. The paper clarifies that ADAS does not introduce a new model, dataset, or deployed system, so these risks are inherited from the underlying models rather than specific to a newly released artifact.
    
\item {\bf Safeguards}
    \item[] Question: Does the paper describe safeguards that have been put in place for responsible release of data or models that have a high risk for misuse (e.g., pre-trained language models, image generators, or scraped datasets)?
    \item[] Answer: \answerNA{} 
    \item[] Justification: ADAS does not release a new model, dataset, or deployed system requiring additional release safeguards. The indirect misuse risks discussed in Section~\ref{sec:discussion} are inherited from the underlying language models rather than from a newly released high-risk artifact.

\item {\bf Licenses for existing assets}
    \item[] Question: Are the creators or original owners of assets (e.g., code, data, models), used in the paper, properly credited and are the license and terms of use explicitly mentioned and properly respected?
    \item[] Answer: \answerYes{} 
    \item[] Justification: We use only existing public benchmarks, models, and evaluation tooling. GSM8K is released under the MIT License; MATH/MATH500 is derived from the MATH dataset repository, released under the MIT License; MBPP is released under CC-BY 4.0; HumanEval is released under the MIT License. We evaluate open-weight Dream-7B-Base, released under Apache-2.0, and LLaDA-8B-Base, released under the MIT License. We use the EleutherAI LM Evaluation Harness, released under the MIT License. We use these assets only for evaluation and cite the corresponding papers or repositories.

\item {\bf New assets}
    \item[] Question: Are new assets introduced in the paper well documented and is the documentation provided alongside the assets?
    \item[] Answer: \answerNA{} 
    \item[] Justification: We do not introduce a new dataset, model, or benchmark. The reproduction code is documented through the supplemental material, while all evaluated datasets and models are existing assets.

\item {\bf Crowdsourcing and research with human subjects}
    \item[] Question: For crowdsourcing experiments and research with human subjects, does the paper include the full text of instructions given to participants and screenshots, if applicable, as well as details about compensation (if any)? 
    \item[] Answer: \answerNA{} 
    \item[] Justification: Our work doesn't involve research with human subjects.

\item {\bf Institutional review board (IRB) approvals or equivalent for research with human subjects}
    \item[] Question: Does the paper describe potential risks incurred by study participants, whether such risks were disclosed to the subjects, and whether Institutional Review Board (IRB) approvals (or an equivalent approval/review based on the requirements of your country or institution) were obtained?
    \item[] Answer: \answerNA{} 
    \item[] Justification: Our work doesn't involve research with human subjects.

\item {\bf Declaration of LLM usage}
    \item[] Question: Does the paper describe the usage of LLMs if it is an important, original, or non-standard component of the core methods in this research? Note that if the LLM is used only for writing, editing, or formatting purposes and does \emph{not} impact the core methodology, scientific rigor, or originality of the research, declaration is not required.
    \item[] Answer: \answerNA{} 
    \item[] Justification: Our usage of LLMs in this work is limited to proofreading of the manuscript.

\end{enumerate}
\end{document}